\documentclass[10pt,twocolumn,letterpaper]{article}

\usepackage{cvpr}              
\usepackage{booktabs}  
\usepackage{multirow}  
\usepackage{adjustbox}  
\usepackage{tabularx,array}
\usepackage{cuted}   
\usepackage{capt-of} 

\newcolumntype{Z}{>{\raggedright\arraybackslash}X}
\newcolumntype{S}[1]{>{\centering\arraybackslash}p{#1}}
\newcolumntype{Y}{>{\centering\arraybackslash}X}

\newcommand{\Idiff}{\mathbf{I}_{\mathrm{diff}}}
\newcommand{\Ihigh}{\mathbf{I}_{\mathrm{high}}}

\newcommand{\I}{\mathbf{I}}

\definecolor{ppurple}{HTML}{A000E0}
\definecolor{ggreen}{HTML}{07A111}
\definecolor{oorange}{HTML}{FF7700}
\definecolor{bblue}{HTML}{0066FF}
\definecolor{yyellow}{HTML}{f6d548}
\definecolor{mmagenta}{HTML}{E03AAE}
\definecolor{cvprblue}{rgb}{0.21,0.49,0.74}
\usepackage[pagebackref,breaklinks,colorlinks,allcolors=cvprblue]{hyperref}

\def\paperID{9392} 
\def\confName{CVPR}
\def\confYear{2026}

\title{UnReflectAnything: RGB-Only Highlight Removal\\by Rendering Synthetic Specular Supervision
}

\author{
Alberto Rota$^{1,*}$,
Mert Kiray$^{2,3}$,
Mert Asim Karaoglu$^{4,2}$,
Patrick Ruhkamp$^{2}$,\\
Elena De Momi$^{1}$,
Nassir Navab$^{2,3}$,
Benjamin Busam$^{2,3}$ \\[6pt]
$^{1}$Politecnico di Milano, Italy \\
$^{2}$Technical University of Munich, Germany \\
$^{3}$Munich Center for Machine Learning (MCML), Germany \\
$^{4}$ImFusion, Germany \\
{\tt\small$^{*}$ alberto1.rota@polimi.it}
}

\begin{document}

\maketitle
\begin{strip}
  \centering
  \includegraphics[width=\textwidth]{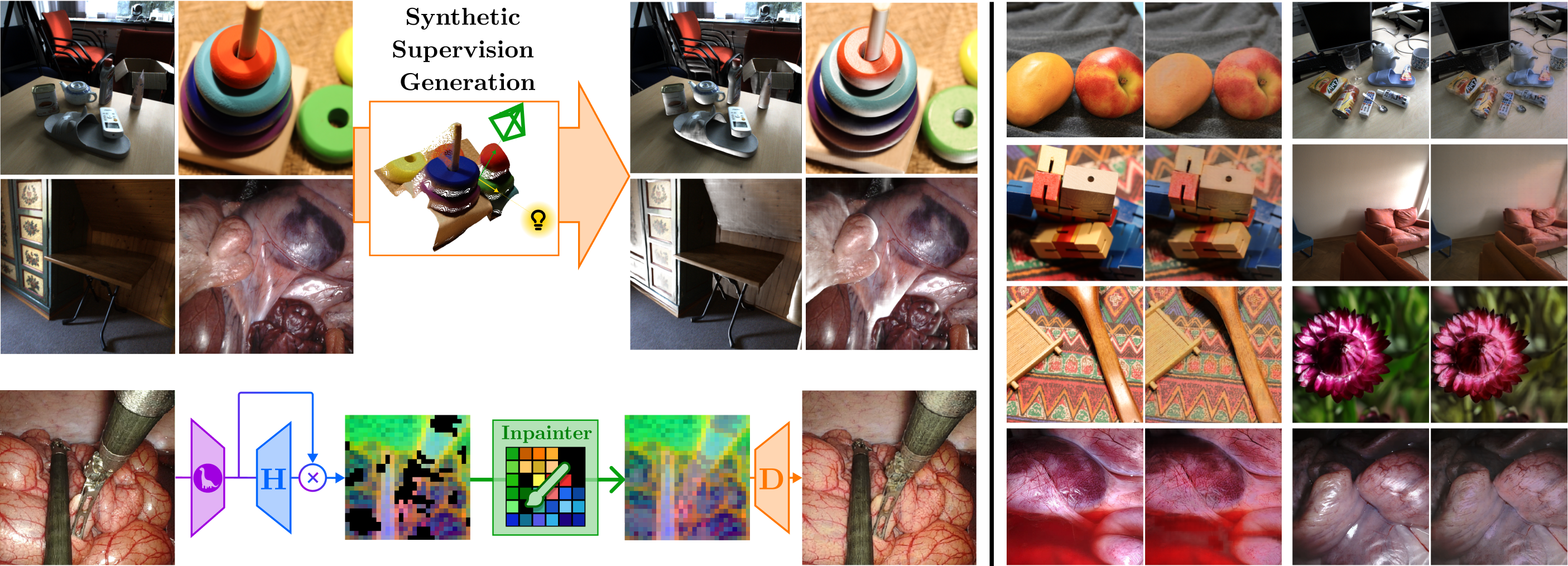}
  \captionof{figure}{\textbf{UnReflectAnything} removes specular highlights from RGB images without paired supervision. A Virtual Highlight Synthesis pipeline renders physically plausible reflections from estimated geometry, providing realistic pseudo-pairs across natural and surgical domains. The model predicts a soft highlight map and inpaints masked DINOv3 tokens, reconstructing reflection-free diffuse images. It generalizes across diverse scenes, achieving faithful highlight suppression and texture recovery.}
  \label{fig:header}
\end{strip}
\begin{abstract}
Specular highlights distort appearance, obscure texture, and hinder geometric reasoning in both natural and surgical imagery. We present UnReflectAnything, an RGB-only framework that removes highlights from a single image by predicting a highlight map together with a reflection-free diffuse reconstruction. The model uses a frozen vision transformer encoder to extract multi-scale features, a lightweight head to localize specular regions, and a token-level inpainting module that restores corrupted feature patches before producing the final diffuse image. To overcome the lack of paired supervision, we introduce a Virtual Highlight Synthesis pipeline that renders physically plausible specularities using monocular geometry, Fresnel-aware shading, and randomized lighting which enables training on arbitrary RGB images with correct geometric structure. UnReflectAnything generalizes across natural and surgical domains where non-Lambertian surfaces and non-uniform lighting create severe highlights and it achieves competitive performance with state-of-the-art results on several benchmarks.
Project Page: \url{https://alberto-rota.github.io/UnReflectAnything/}
\end{abstract}    

\section{Introduction}

Specular highlights arise from mirror-like reflections on non-lambertian surfaces, producing saturated, view-dependent artifacts that obscure scene content and corrupt downstream perception tasks such as segmentation, correspondence, and photometric inference~\cite{Fu2021CVPR,Wu2022TMM}. In natural images, these highlights distort color fidelity and degrade restoration or relighting pipelines. In endoscopic scenes, bodily fluids, moist tissue and strong non-uniform lighting introduce intense specularities that occlude anatomy, bias texture cues, and hinder navigation and decision-making~\cite{Daher2023MedIA,Zhang2023PMB,karaoglu2024ride,karaoglu2025litetracker}. Such artifacts critically impair depth estimation, optical flow, and stereo reconstruction in surgical robotics~\cite{Daher2023MedIA}, where accurate scene understanding is essential and visual ambiguity can carry clinical consequences.

Removing specular highlights from a \emph{single RGB image} is fundamentally ill-posed: the specular component is often saturated, spatially sparse, and entangled with scene geometry, material properties, and illumination. While polarization cameras can physically disentangle diffuse and specular reflections, their specialized hardware limits practicality in most consumer and surgical systems~\cite{guo2018single}.

To this end, prior works have extensively explored both classical and learning-based approaches for specular highlight removal using only RGB images. Classical methods rely on color priors~\cite{Tan2005PAMI}, color ratios~\cite{Shen2013AO}, or dichromatic reflectance models~\cite{Shafer1985Dichromatic} to estimate diffuse components. However, these models often break down under real-world conditions involving complex materials or uncontrolled illumination. More recently, deep learning methods~\cite{Fu2021CVPR,Wu2022TMM} have leveraged data-driven priors to learn the separation of diffuse and specular layers directly from examples. Yet, the scarcity of paired supervision and the domain shift between synthetic and medical imagery limit their generalization. In practice, these networks tend to over-smooth texture or introduce hue distortions near highlight boundaries, reducing fidelity in scenes that demand accurate appearance preservation.

We introduce \textbf{UnReflectAnything} (Fig.~\ref{fig:header}), an RGB-only framework for single-image specular highlight removal across natural and surgical imagery.
Our main contributions are:

\begin{itemize}
\item \textbf{Virtual highlight synthesis.} We render realistic specularities from monocular geometry using Fresnel-aware shading and randomized lighting, enabling supervision without paired diffuse-only data.

\item \textbf{Token-space diffuse inpainting.} A transformer inpainter reconstructs masked DINOv3 patch tokens directly in feature space, restoring diffuse appearance with global context.  

\item \textbf{Hybrid geometry and pixel-level supervision.} A unified training scheme couples synthetic highlight rendering with token- and image-space losses, enforcing seamless boundary consistency and cross-domain robustness.  
\end{itemize}    
\section{Related Work}
Specular highlight removal is a long-standing challenge in computer vision.
Early approaches relied on physical priors to suppress specular reflections.
Tan \etal~\cite{Tan2005PAMI} and Shen \etal~\cite{Shen2013AO} leveraged color constancy and color ratio cues to decompose diffuse and specular components, while the dichromatic reflection model~\cite{Shafer1985Dichromatic} offered a principled formulation under simplified reflectance and illumination assumptions.
Hardware-based techniques such as multi-flash imaging~\cite{Feris2006JBCS} or polarization cameras~\cite{lee2022reduction} enable more explicit separation of reflection components, yet their specialized setups are often impractical for consumer or surgical environments.

Learning-based methods have rapidly advanced specular highlight removal through large-scale data and task-coupled architectures.
HighlightNet~\cite{Fu2021CVPR} established the first large-scale dataset of synthetic and real images with joint highlight detection and removal supervision.
SpecularityNet~\cite{Wu2022TMM} leveraged cross-polarization capture and adversarial training.
To mitigate the need for paired data, MG-CycleGAN~\cite{Hu2022PRL} introduced soft highlight masks and cycle-consistent adversarial training, enabling unpaired learning.
More recently, DHAN-SHR~\cite{Guo2024ACMMM} employed local–global hybrid transformers to achieve strong diffuse–specular decomposition without explicit mask supervision.
Diffusion-based approaches have also emerged for specular suppression; StableDelight~\cite{StableDelight2025}, inspired by StableNormal~\cite{ye2024stablenormal}, leverages a generative diffusion framework trained on large-scale synthetic and real data to remove reflections and recover surface details.

While these approaches have shown strong performance on natural images, their adaptation to medical imagery presents unique challenges.
Endoscopic scenes often contain broad, high-intensity reflections caused by moist tissue and dynamic lighting, making standard RGB-based solutions less reliable.
To handle such domain-specific characteristics, several works have been proposed for endoscopic applications.
Seminal efforts explored robust PCA-based decomposition~\cite{Li2019TMI}, partial convolutions for specularity inpainting~\cite{Zhang2023PMB}, and attention-driven architectures such as Endo-STTN~\cite{Daher2023MedIA}, which employs temporal transformers to propagate texture across neighboring frames and enhance realism in surgical videos.
However, Endo-STTN relies on binary highlight masks as explicit input queries, which are often unavailable in practice.
Despite recent progress, paired surgical datasets remain scarce and existing models still struggle with severe saturation and complex illumination commonly encountered in endoscopic scenes.

\begin{figure*}
    \centering
    \includegraphics[width=\linewidth]{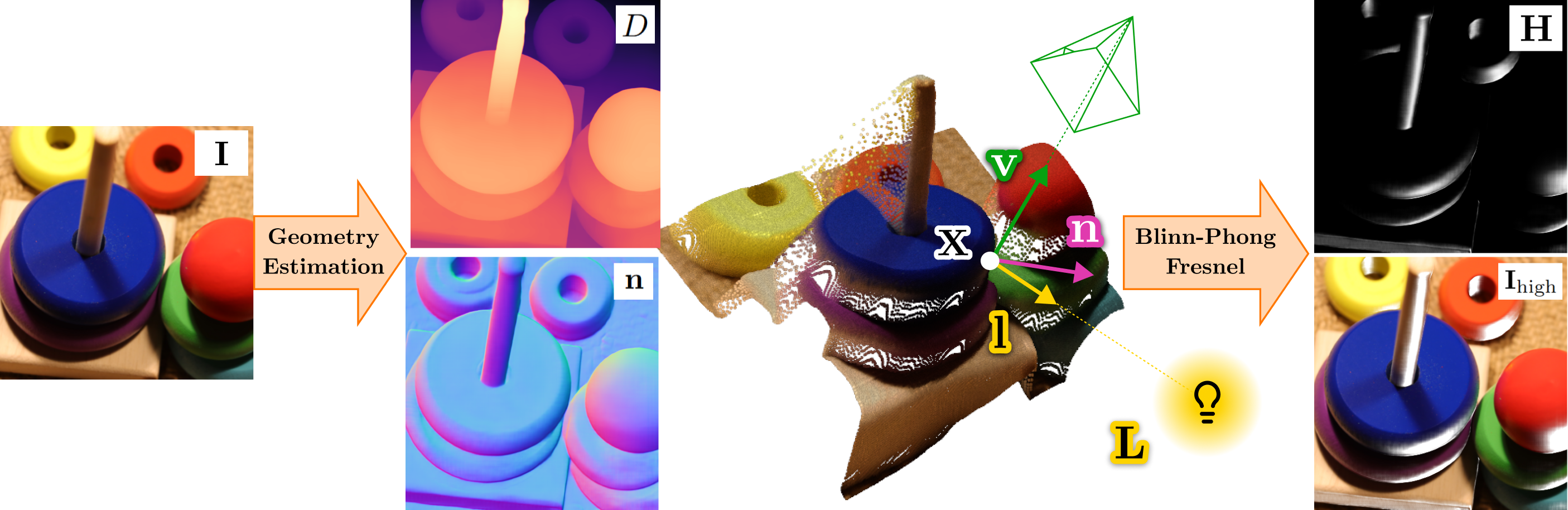}
     \caption{\textbf{Synthetic Highlight Generation Pipeline from any image.} Given a single RGB image (left), a per-pixel depth and surface normals is first estimated using a monocular geometry network. The recovered geometry defines a 3D point cloud $\mathbf{X}$ with associated \textcolor{mmagenta}{surface normals $\mathbf{n}$} and \textcolor{ggreen}{view directions $\mathbf{v}$}. The light source position $\mathbf{L}$ is sampled in camera coordinates, producing local \textcolor{yyellow}{illumination vectors $\mathbf{l}$}. These geometric quantities drive a physically based Blinn–Phong \cite{blinnphong} rendering mode generating a synthetic highlight intensity map that is photometrically consistent with the inferred scene structure. The highlight is finally composited with the input RGB.}
    \label{fig:highlights}
\end{figure*}

Polarization data inherently disentangle specular and diffuse components, providing a powerful physical prior for both training and inference conditioning in learning-based highlight removal.
SHM-GAN~\cite{Anwer2023NC} employed polarized supervision to enable RGB-only inference, while PolarAnything~\cite{Zhang2025ICCV} (PA) synthesized polarimetric cues (DoLP/AoLP) directly from RGB inputs, enabling polar-aware processing without specialized sensors.
PolarFree~\cite{Yao2025CVPR} (PF) leveraged diffusion priors with large-scale RGB–polarization pairs to enhance glare suppression, although it still depends on polarized inputs during inference.
While effective, these approaches introduce additional sensing or calibration requirements that limit their practicality in compact or sensor-constrained imaging systems.

To our knowledge, \textit{UnReflectAnything} is the first work to combine monocular geometry, Fresnel-aware specular rendering, and randomized lighting to generate physically plausible paired supervision for RGB-only highlight removal across both natural and surgical imagery.
    
\begin{figure*}
    \centering
    \includegraphics[width=\linewidth]{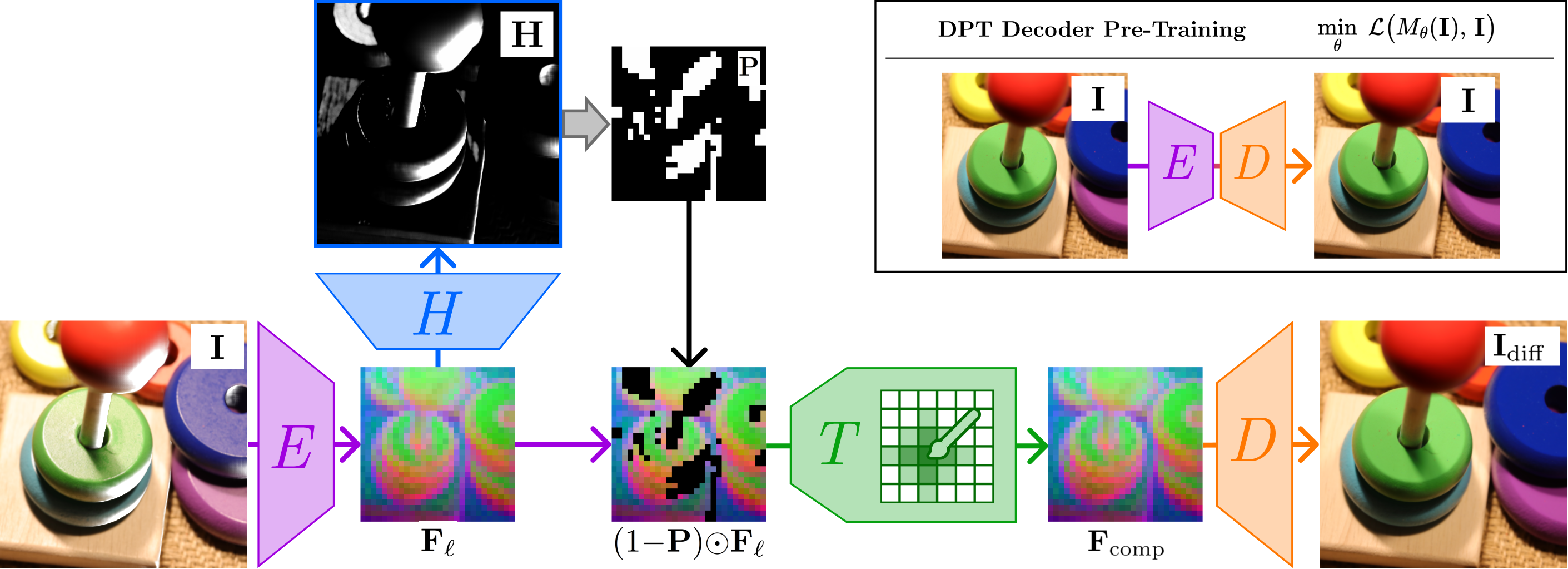}
    \caption{\textbf{UnReflectAnything model architecture.}
    A pretrained \textcolor{ppurple}{DINOv3 encoder backbone $\mathit{E}$} extracts a hierarchy of multi-scale patch features from the input image (only the last feature map in the hierarchy is shown for clarity). A DPT-inspired \textcolor{bblue}{highlight predictor $\mathit{H}$} produces a soft, pixel-level highlight map, serving as a mask on the feature maps. A lightweight ViT-based \textcolor{ggreen}{Token Inpainter $\mathit{T}$} operates on these masked features, learning to reconstruct the underlying diffuse features in place of those corrupted by highlights. An \textcolor{oorange}{RGB DPT decoder $\mathit{D}$} transforms the inpainted feature maps into a reflection-free diffuse RGB image. This decoder is pre-trained in an autoencoderfashion to reconstruct the input RGB image from frozen DINOv3 features by minimizing the pixel-wise reconstruction loss $\mathcal{L}(M_\theta(\I), \I)$.}
    \label{fig:model}
\end{figure*}
\section{Methodology}
We integrate physically-grounded virtual highlight synthesis with token-space inpainting to recover reflection-free images from a single RGB view. The pipeline couples geometric rendering, highlight localization, and feature-level reconstruction into a unified framework.
\subsection{Virtual Highlight Synthesis}
\label{sec:highlightsynthesis}
Accurate 3D geometry estimates enable the synthesis of realistic highlights on any RGB image, supplying effective supervision during training.
Given an input linear RGB image $\mathbf{I}\in[0,1]^{H\times W\times 3}$, we generate physically--plausible synthetic specular highlights by (i) inferring scene geometry from a single view, (ii) sampling a point light in camera space, and (iii) rendering a Blinn--Phong specular lobe with Fresnel modulation. The resulting highlights serve as a supervision signal during training. The virtual highlight synthesis pipeline is illustrated in Fig.~\ref{fig:highlights}.

\medskip\noindent\textbf{Monocular Geometry.}
We estimate metric depth $D \in \mathbb{R_+}^{H\times W}$, surface normals $\mathbf{n} \in  \mathbb{R}^{H\times W \times 3}$, and intrinsics $\mathbf{K} \in \mathbb{R}^{3\times3}$ using an off-the-shelf method.
For each pixel in homogeneous coordinates $p=(u,v,1)^{\top}$, we back--project into its correspo\textbf{}nding 3D camera coordinates $\mathbf{X}$ with 
$
\mathbf{X} = D(p)\,\mathbf{K}^{-1}p. 
$
The position of a point light $\mathbf{L} \in \mathbb{R}^3$ is sampled in 3D camera coordinates. We define the normalized view and light directions as
\begin{equation}
\mathbf{v} = \frac{\mathbf{X}}{\|\mathbf{X}\|},\qquad
\mathbf{l} = \frac{\mathbf{L}-\mathbf{X}}{\|\mathbf{L}-\mathbf{X}\|}.
\end{equation}

\medskip\noindent\textbf{Specular Core.}
Let $\mathbf{h}=\frac{\mathbf{l}+\mathbf{v}}{\|\mathbf{l}+\mathbf{v}\|}$ be the half-vector between the light and view direction. For each point $\mathbf{x}$, we compute  the Schlick-Fresnel \cite{schlick} reflection coefficient
\begin{equation}
    R = R_{0}+(1-R_0)\left(1-\mathbf{v}\cdot \mathbf{h}\right)^{5},
\end{equation}
where $R_0$ is the approximation of the Fresnel reflectance at normal incidence. We employ the Blinn-Phong \cite{blinnphong} reflection model to compute the per-pixel specular highlight intensity
\begin{equation}
\mathbf{H} = K_H R \left(\mathbf{n}\cdot \mathbf{h}\right)^{S},
\end{equation}
where $K_H>0$ controls the global highlight intensity and $S$ quantifies the surface shininess.
At training time $K_H$,  $S$ and the light position $\textbf{L}$ are sampled from empirically tuned uniform distributions that produce perceptually realistic highlight patterns, thereby increasing the heterogeneity of the supervision signal.

\medskip\noindent\textbf{RGB Compositing.}
To obtain the final highlight-augmented image, the rendered specular lobe is
\emph{alpha-composited} onto the raw RGB input. Specifically, we treat the
highlight intensity $H$ as an additive alpha mask scaled by a global
scaling factor $K_H$, and blend it with the original pixel $\mathbf{I}$:
\begin{equation}
\Ihigh
= 
(1 - \mathbf{H})\,\mathbf{I}
+ \mathbf{H}\,\big(\mathbf{I} + K_H\,\mathbf{1}_3\big).
\end{equation}

A critical aspect of training is that the original input $\mathbf{I}$ is \emph{not} guaranteed to be highlight-free; on the contrary most images natively exhibit real specularities, especially endoscopy images. We denote these naturally occurring reflections as \emph{dataset highlights}, hence specular regions already present in the raw data and not introduced synthetically. To differentiate them from our \emph{synthetic highlights}, we detect \textit{dataset highlights} via thresholding with a high luminance cutoff, $\tau_L$, on the raw RGB images.
Since supervising these pixels would mislead the network into interpreting saturated highlight regions as diffuse ‘white’ surfaces, we explicitly exclude all dataset-highlight pixels from supervision during training.

\subsection{Model Architecture}
\label{sec:model}
Our model $M$ inputs an RGB image $\mathbf{I} \in [0,1]^{H \times W \times 3}$ and jointly predicts two outputs: a reflection-free diffuse-only image $\Idiff \in [0,1]^{H \times W \times 3}$ and a score map for highlights $\Ihigh \in [0,1]^{H \times W}$ that encodes the probability of specular regions:
$
    (\Idiff, \Ihigh) = M(\mathbf{I}).
$

$M$ consists of four main components, as we illustrate in Fig.~\ref{fig:model}: an encoder $E$ that extracts hierarchical patch-level features from the input image, a highlight predictor head $H$ that regresses a continuous highlight probability map, a token inpainter $T$ that reconstructs masked feature regions corresponding to predicted highlights, and an RGB decoder $D$ that synthesizes the reflection-free diffuse image from the inpainted features. The overall process can be expressed formally as:
$
    \Ihigh = H\big(E(\mathbf{I})\big)$ and 
    $\Idiff = D\big(T(E(\mathbf{I}), \Ihigh)\big)
$.
The highlight predictor guides the inpainter by localizing overexposed or specular areas, enabling the model to recover the underlying diffuse appearance before decoding it into the final RGB output.

\medskip\noindent\textbf{Feature Extraction.}
We use a frozen DINOv3-Large~\cite{simeoni2025dinov3} Vision Transformer as encoder $E$, which maps the input image $\mathbf{I}$ to patch-level feature tokens at four depths:
\begin{equation}
\mathcal{F} = \{\mathbf{F}_1,\mathbf{F}_2,\mathbf{F}_3,\mathbf{F}_4\} = E(\mathbf{I}),
\qquad \mathbf{F}_\ell \in \mathbb{R}^{N \times C}.
\end{equation}

These multi-scale tokens provide the hierarchical representations later consumed by the decoders for pixel-space reconstruction.

\medskip\noindent\textbf{Highlight Prediction.}
A lightweight decoder $H$ takes the multi-scale token set $\mathcal{F}$ and predicts a single-channel highlight intensity map $\Ihigh \in [0,1]^{H \times W}$. This map identifies regions affected by specular reflections, providing a guidance signal for subsequent token inpainting.

\begin{figure}
    \centering
    \includegraphics[width=\linewidth]{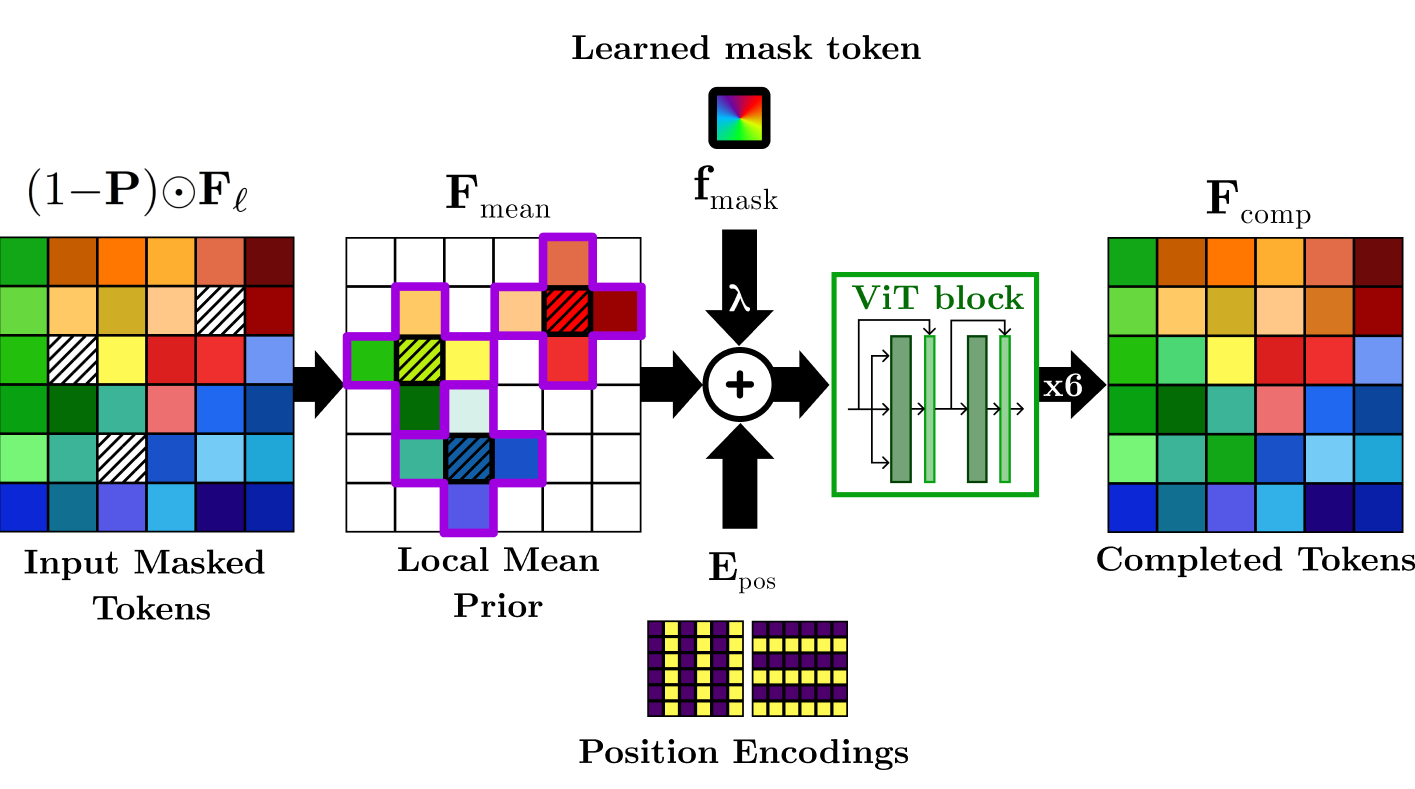}
    \caption{\textbf{Patch token inpainting logic.} A local neighborhood (purple borders) for each token  to be inpainted is used to compute the local mean priors. This mean priors are summed with the Positional Embeddings and a learned mask token and fed into a sequence of transformer blocks which refine the tokens to the final feature.}
    \label{fig:tokeninpaint}
\end{figure}

\medskip\noindent\textbf{Patch-Token Diffuse Inpainting.}
The token inpainter $T$ (Fig.~\ref{fig:tokeninpaint}) reconstructs feature tokens corresponding to highlight-contaminated regions. Given the encoder features $\mathbf{F}_\ell \in \mathbb{R}^{N \times C}$ and a patch mask $\mathbf{P} \in [0,1]^{N}$, $T$ operates entirely in token space to infer the missing information and produce a coherent diffuse representation. Each of the following operations are performed separately at each $\ell$ layer, therefore we drop the $\ell$ notation for the rest of the section.  $\mathbf{P}$ is obtained by spatially downsampling $\Ihigh$ to the encoder patch resolution using average pooling and thresholding the mean highlight intensity per patch.

Patch tokens to be inpainted are first substituted with a learnable \emph{mask token} $\mathbf{f}_{\mathrm{mask}}$, blended with a local mean prior $\mathbf{F}_{\mathrm{mean}}$ computed from visible patch neighbors via depthwise convolution and enriched by adding fixed 2D positional encodings $\mathbf{E}_{\mathrm{pos}}$:
\begin{equation}
    \mathbf{F}_{\mathrm{seed}} = \mathbf{P} \odot \big[\lambda\,\mathbf{f}_{\mathrm{mask}} + (1-\lambda)\,\mathbf{F}_{\mathrm{mean}}\big] + (1-\mathbf{P}) \odot \mathbf{F} + \mathbf{E}_{\mathrm{pos}},
\end{equation}
where $\lambda$ is the local mean prior coefficient. 
This \emph{seed} sequence is then refined through a stack of six vision transformer layers performing self-attention over both visible and masked tokens. The final completed representation merges the inpainted and original visible tokens:
\begin{equation}
\mathbf{F}_{\mathrm{comp}} = \mathbf{P} \odot \operatorname{ViT}(\mathbf{F}_{\mathrm{seed}}) + (1 - \mathbf{P}) \odot \mathbf{F}.
\end{equation}

The resulting multi-scale set $\mathcal{F}_{\mathrm{comp}} = \{\mathbf{F}_{\mathrm{comp}}\,\}$ encodes highlight-free features that preserve both spatial structure and semantic coherence, and is subsequently decoded by $D$ to reconstruct the diffuse image $\Idiff$. 


\medskip\noindent\textbf{Diffuse Image Reconstruction.}
We utilize a lighweight decoder to reconstruct diffuse RGB images, $\Idiff$, from completed multi-scale tokens, $\mathcal{F}_{\mathrm{comp}}$.

\begin{figure}[t]
    \centering
    \includegraphics[width=\linewidth]{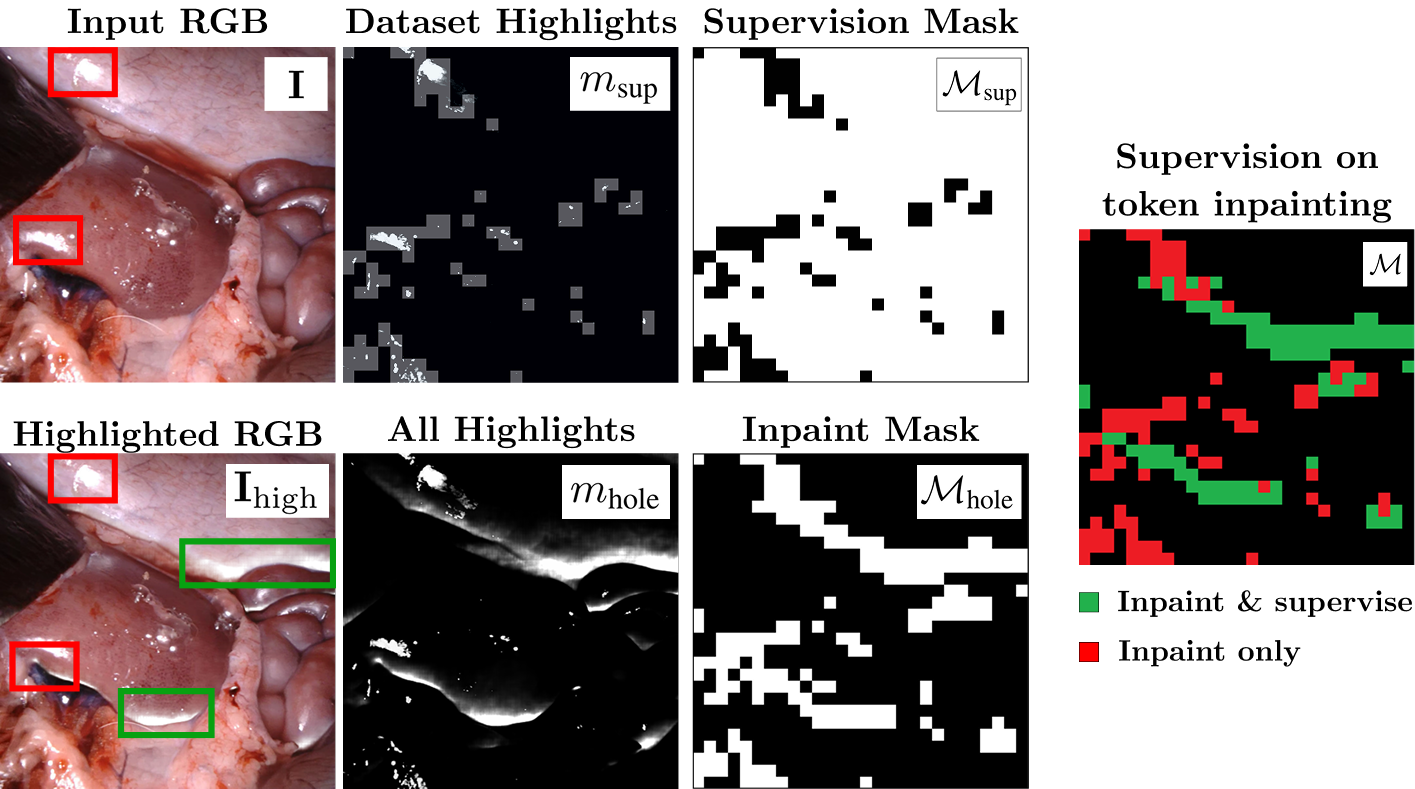}
    \caption{\textbf{Supervision masks for the token inpainting module.} Reliable supervision excludes \textit{dataset highlight} regions (red). The inpainter must, however, learn to  complete \emph{all} highlight regions, including both synthetic (green) and dataset highlights.}
    \label{fig:supervisionmasks}
\end{figure}

\begin{figure*}
    \centering
    \includegraphics[width=\linewidth]{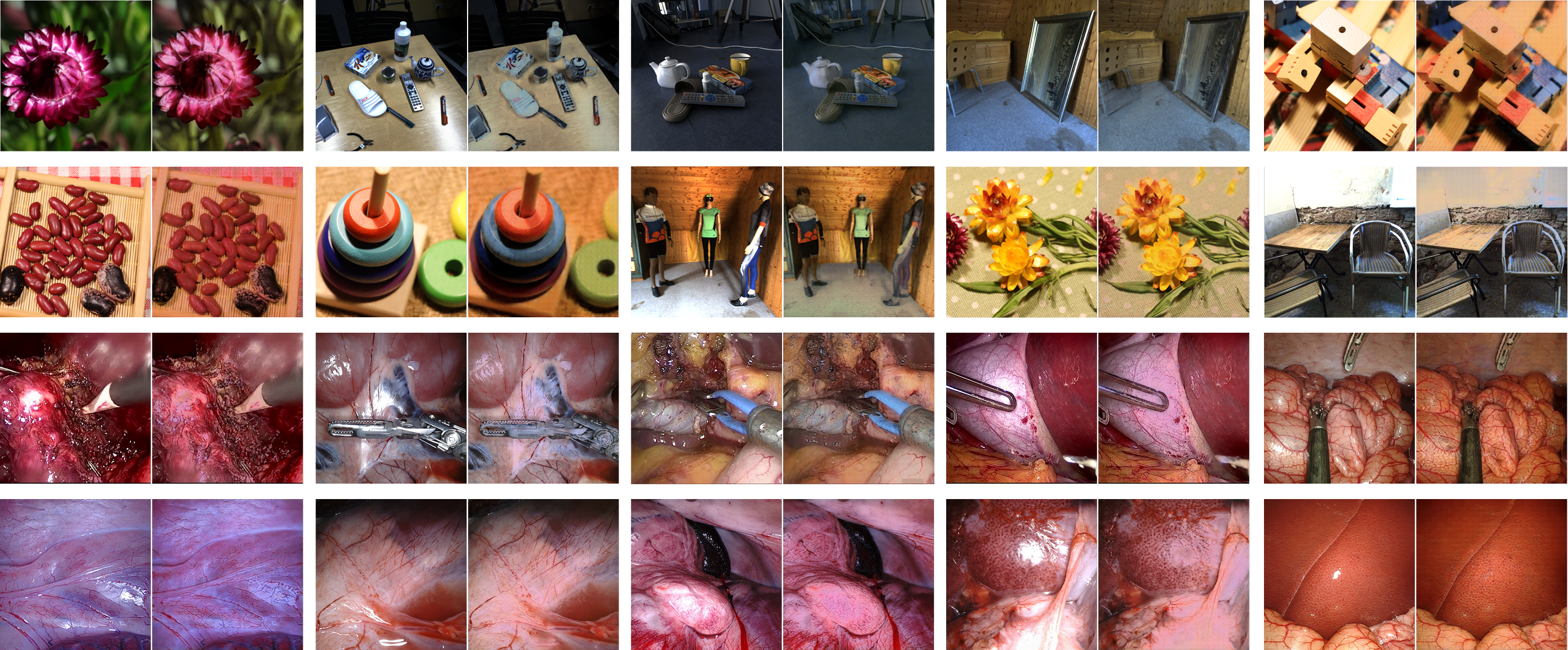}
    \caption{\textbf{Qualitative examples} for pairs of raw images with highlights (left) and their \textit{UnReflectAnything}-processed counterpart (right). Our framork consistently removes, inpaints or attenuates specular highlights from images in several different domains.}
    \label{fig:qualitativeresults}
\end{figure*}
\subsection{Supervision}
\label{sec:supervision}

Our model is trained under a hybrid supervision scheme combining synthetic highlight annotations, pixel–space signals, and token–level objectives. The primary source of supervision is the \emph{Synthetic Highlight Generation} pipeline, which enables pairing any RGB image $\I$ with a soft highlight mask $\mathbf{H}$ and its corresponding synthetically highlighted image $\Ihigh$.

Training relies on two pixel masks:  
(i) a \emph{supervision mask} $m_{\mathrm{sup}} \in \{0,1\}^{H\times W}$ identifying pixels where ground truth is trustworthy (\textit{i.e.}, regions free of inherent \emph{dataset highlights}), and  
(ii) a \emph{hole mask} $m_{\mathrm{hole}} \in \{0,1\}^{H\times W}$ denoting all regions to be inpainted, formed as the union of synthetic and dataset highlights.  

For token–space supervision, both masks are downsampled to the patch grid (stride $p$), yielding the binary patch sets  
$\mathcal{M}_{\mathrm{sup}} \subset \{1,\dots,N\}$ and $\mathcal{M}_{\mathrm{hole}} \subset \{1,\dots,N\}$, which indicate patches eligible for reliable supervision and patches requiring inpainting, respectively.

\medskip\noindent\textbf{Highlight Supervision.}
The highlight–prediction head $H$ is trained using a weighted sum of soft Dice, L1, and Total Variation losses as 
$
\label{eq:hl_short}
\mathcal{L}_{\mathrm{H}}
=
w_{\mathrm{dice}}\mathcal{L}_{\mathrm{dice}}
+
w_{\mathrm{L1}}\mathcal{L}_{\mathrm{L1}}
+
w_{\mathrm{TV}}\mathcal{L}_{\mathrm{TV}}
$.

\medskip\noindent\textbf{Token–Space Inpainting Supervision.}
The token inpainter predicts cleaned diffuse-only features for all highlight–affected patches, including both synthetically inserted and dataset-originating highlights. Because dataset highlights correspond to saturated or clipped intensities, their tokens do not constitute meaningful targets. Thus, we restrict supervision to the intersection 
$\mathcal{M}=\mathcal{M}_{\mathrm{hole}} \cap \mathcal{M}_{\mathrm{sup}}$, which identifies the subset of inpainted patches equipped with reliable ground truth. Fig.~\ref{fig:supervisionmasks} provides a visual interpretation.
For each patch $i\in\mathcal{M}$, the ground truth token is taken from the original clean image,
$\mathbf{F}^*_{\ell} = E(\I)$, before the synthetic highlight pipeline is applied. The inpainting loss utilizes cosine similarity and L1 loss, controlled by $\alpha$:
\begin{equation}
\mathcal{L}_{\mathrm{inp}}
=
\frac{1}{|\mathcal{M}|}
\sum_{i\in\mathcal{M}}
\Big[
\alpha\,\|\mathbf{F}^*_{i}-\mathbf{F}_{i}\|_{1}
+
(1-\alpha)\!\left(
1-\mathbf{F}_i^{\!*}\,\mathbf{F}_i^{\top}
\right).
\end{equation}

\medskip\noindent\textbf{Decoder Pre–Training.}
Before inpainting is introduced, the RGB decoder $D$ is \emph{pre–trained} with a frozen DINOv3 encoder $E$ in an auto-encoder fashion. The objective encourages the decoder to faithfully reconstruct the input RGB image from its DINOv3 features, with minimization of an L1 and SSIM term for RGB reconstruction:
\begin{equation}
\mathcal{L}_{\mathrm{AE}}
=
\|D(E(\mathbf{I})) - \mathbf{I}\|_{1}
+
\big(1-\mathrm{SSIM}(D(E(\mathbf{I})),\,\mathbf{I})\big).
\end{equation}
This initialization ensures that $D$ provides a stable and semantically meaningful feature-to-image mapping before the inpainting module is trained.
\begin{table*}
\centering
\footnotesize
\caption{Comparison across datasets that provide paired diffuse-only ground truth. The best score in each metric is highlighted in \textbf{bold}.}
\label{tab:withgt}
\begin{tabularx}{\linewidth}{@{}l|YYY|YYY|YYY@{}}
\toprule
\textit{} & \multicolumn{3}{c|}{PSD} & \multicolumn{3}{c|}{SHIQ} & \multicolumn{3}{c}{SSHR} \\
\cmidrule{2-10}
\cmidrule{2-10}
 \textit{Method} & $\mathrm{MSE}_{m}$ $\downarrow$ & PSNR $\uparrow$ & SSIM $\uparrow$ & $\mathrm{MSE}_{m}$ $\downarrow$ & PSNR $\uparrow$ & SSIM $\uparrow$ & $\mathrm{MSE}_{m}$ $\downarrow$ & PSNR $\uparrow$ & SSIM $\uparrow$ \\
\midrule
DHAN-SHR~\cite{guo2024dual}                       & 0.006           & \textbf{24.863} & 0.868          & 0.001             & \textbf{33.815} & 0.982          & \textbf{0.001} & \textbf{37.318} & \textbf{0.971}  \\
PA+PF~\cite{Yao2025CVPR}~\cite{Zhang2025ICCV} & 0.050           & 14.101          & 0.645          & 0.089             & 12.009          & 0.604          & 0.024          & 18.102          & 0.432          \\
SpecularityNet~\cite{Wu2022TMM}                   & 0.007           & 23.675          & 0.839          & 0.005             & 27.386          & 0.961          & 0.002          & 31.628          & 0.952          \\
Fu \textit{et al.}~\cite{fu2023towards}           & 0.007           & 23.786          & 0.838          & \textbf{0.001}    & 32.885          & 0.978          & 0.001          & 35.857          & 0.963          \\
Li \textit{et al.}~\cite{li2025two}               & 0.019           & 18.819          & 0.716          & 0.007             & 24.351          & 0.919          & 0.004          & 26.898          & 0.909          \\
EndoSTTN~\cite{daher2023temporal}                 & 0.009           & 22.951          & 0.844          & 0.005             & 27.203          & 0.955          & 0.003          & 30.316          & 0.951          \\
StableDelight~\cite{ye2024stablenormal}           & 0.007           & 23.384          & 0.838          & 0.007             & 22.844          & 0.921          & 0.001          & 32.838          & 0.886          \\
\midrule
OURS                                              & \textbf{0.004}  & 17.230          & \textbf{0.911} & 0.003             & 17.610          & \textbf{0.988} & 0.002         & 28.415           & \textbf{0.971}          \\

\bottomrule
\end{tabularx}
\end{table*}

\medskip\noindent\textbf{Decoder Fine-Tuning.}
After pre–training, the RGB decoder $D$ is fine–tuned to restore seamless boundaries along inpainted regions, suppress residual highlights in the diffuse output $\Idiff$, and re–align feature statistics with the expected input distribution.  
To enforce smooth transitions at inpainting borders, \textit{i-e.} seams, we define a thin boundary ring 
$
r = \mathrm{dilate}(m_{\mathrm{hole}}) - m_{\mathrm{hole}}
$ 
and constrain discontinuities only within this region:
\begin{equation}
\mathcal{L}_{\mathrm{seam}}
= 
\|(\Idiff-\I)\odot r\|_{1}
+
\lambda_{g}\|\nabla \Idiff - \nabla \I\|_{1,\;r}.
\end{equation}
This term promotes color and gradient continuity across the inpainting boundary.

To prevent the diffuse decoder from reintroducing specular peaks, we additionally penalize overly bright pixels in $\Idiff$ using a smooth Charbonnier loss.  
Let $B=\tfrac{1}{3}(I_R+I_G+I_B)$ and $m_{\mathrm{hl}}=[B>\tau_m]$,
we define:
\begin{equation}
\mathcal{L}_{\mathrm{spec}}
=
\frac{1}{|m_{\mathrm{hl}}|}
\sum_{p\in m_{\mathrm{hl}}}
\sqrt{(B(p)-\tau)^2+\varepsilon^2}.
\end{equation}

We reintroduce the same RGB reconstruction loss used during decoder pre–training, this time comparing the predicted diffuse image $\Idiff$ with its reference $\Idiff^{*}$ while excluding dataset highlights via the supervision mask $m_{\mathrm{sup}}$:
\begin{equation}
\mathcal{L}_{\mathrm{RGB}}
=
\|\Idiff-\Idiff^{*}\|_{1}
+
\big(1-\mathrm{SSIM}(\Idiff,\Idiff^{*})\big).
\label{eq:rgbloss}
\end{equation}

\section{Experiments}
Experiments are performed on 1x NVIDIA A100 GPU (80~GB), and trained with a batch size of 32 for 50 epochs; we set an initial learning rate of $5\times10^{-4}$, decaying linearly every 10 epochs.
We use MoGe--2~\cite{wang2025moge} as the metric depth estimation model.
For the token in-painter, we use sequence of ViT-layers.
For both highlight prediction head, $H$ and diffuse RGB decoder $D$, we utilize a Dense Prediction Transformer (DPT)-based architecture~\cite{ranftl2021vision}.

We empirically set the approximation of the Fresnel reflectance at normal incidence, $R_0$, to $0.04$; the luminance threshold, $\tau_L$, to $0.95$; the local mean prior coefficient, $\lambda$, to $0.5$; token-inpainting coefficient, $\alpha$, to $0.25$; and the mask threshold decoder fine-tuning, $\tau_m$, to $0.85$ with its corresponding stability bias, $\epsilon$, to $10^{-6}$.
Cf. Sup. Mat. for details on loss weights and scheduling.






\begin{table}
\centering
\scriptsize
\caption{Comparison across datasets that do not provide paired diffuse-only ground truth. The best score in each metric is highlighted in \textbf{bold}. For all metrics, lower is better.}
\label{tab:nogt}
\begin{tabularx}{\linewidth}{@{}l|Y@{\hspace{1.5mm}}Y|Y@{\hspace{1.5mm}}Y|Y@{\hspace{1.5mm}}Y}

\midrule
\multicolumn{7}{c}{%
    {\fontsize{6.5pt}{7.5pt}\selectfont\textsc{\textbf{Natural Domain}}}%
}\\
\midrule{}
\textit{} & \multicolumn{2}{c|}{CroMo} & \multicolumn{2}{c|}{HouseCat6D} & \multicolumn{2}{c}{SCRREAM} \\
\cmidrule(r){2-3} \cmidrule(r){4-5} \cmidrule(r){6-7}
\textit{Method} & LSR & NIQE  & LSR & NIQE  & LSR & NIQE \\
\midrule
DHAN-SHR~\cite{guo2024dual}                       & 0.063              & 0.084             & 0.224              & 0.142              & 0.073              & 0.075 \\
PA+PF~\cite{Yao2025CVPR}~\cite{Zhang2025ICCV} & 0.160              & 0.103             &0.280               & 0.167              & 0.206             & 0.093 \\
SpecularityNet~\cite{Wu2022TMM}                   & 0.179              & 0.070             & 0.192              & 0.141              & 0.109              & 0.070 \\
Fu \textit{et al.}~\cite{fu2023towards}           & 0.129              & 0.068             & 0.241              & 0.132              & 0.147              & 0.059 \\
Li \textit{et al.}~\cite{li2025two}               & 0.113              & 0.088             & 0.184              & 0.143              & 0.088              & 0.085 \\
EndoSTTN~\cite{daher2023temporal}                 & 0.028              & 0.095             & 0.078              & 0.149              & 0.039              & 0.078 \\
StableDelight~\cite{ye2024stablenormal}           & 0.176              & 0.073             & 0.318              & 0.142              & 0.270              & 0.078 \\
\midrule
OURS                                              & \textbf{0.012}     & \textbf{0.061}    & \textbf{0.033}     & \textbf{0.120}     & \textbf{0.002}     & \textbf{0.056} \\
\midrule
\midrule
\multicolumn{7}{c}{%
    {\fontsize{6.5pt}{7.5pt}\selectfont\textsc{\textbf{Surgical Domain}}}%
}\\
\midrule
\textit{} & \multicolumn{2}{c|}{Cholec80} & \multicolumn{2}{c|}{SCARED} & \multicolumn{2}{c}{StereoMIS-Tr} \\
\cmidrule(r){2-3} \cmidrule(r){4-5} \cmidrule(r){6-7}
\textit{Method}                                    & LSR                & NIQE              & LSR                & NIQE               & LSR                & NIQE \\
\midrule
DHAN-SHR~\cite{guo2024dual}                        & 0.115              & 0.101             & 0.114              & 0.131              & 0.077              & 0.165 \\
PA+PF~\cite{Yao2025CVPR}~\cite{Zhang2025ICCV}  & 0.111              & 0.161             & 0.110              & 0.162              & 0.151              & 0.191 \\
SpecularityNet~\cite{Wu2022TMM}                    & 0.150              & 0.099             & 0.103              & 0.116              & 0.112              & 0.115 \\
Fu \textit{et al.}~\cite{fu2023towards}            & 0.157              & 0.072             & 0.147              & \textbf{0.098}     & 0.111              & 0.115 \\
Li \textit{et al.}~\cite{li2025two}                & 0.035              & 0.108             & 0.021              & 0.151              & 0.038              & 0.134 \\
EndoSTTN~\cite{daher2023temporal}                  & 0.021              & 0.135             & 0.020              & 0.175              & 0.052              & 0.163 \\
StableDelight~\cite{ye2024stablenormal}            & 0.316              & \textbf{0.046}    & 0.262              & 0.103              & 0.278              & \textbf{0.110} \\
\midrule
OURS                                               & \textbf{0.002}     & 0.100             & \textbf{0.011}     & 0.144              & \textbf{0.022}     & 0.131 \\
\bottomrule
\end{tabularx}
\end{table}

\subsection{Datasets}
\label{sec:datasets}
In order to capture a wide range of domains, we utilize a combination of indoor (SCRREAM~\cite{jung2024scrream}, HouseCat6D~\cite{jung2024housecat6d}), outdoor (CroMo~\cite{verdie2022cromo}), and endoscopic (SCARED~\cite{allan2021scared}, Cholec80~\cite{twinanda2016cholec}) datasets.

For evaluation purposes, we group datasets into two categories based on the availability of diffuse ground truth. 
\textit{Diffuse-referenced datasets} provide paired diffuse images $\Idiff^\star$ for each sample, including PSD~\cite{wu2021psd}, SHIQ~\cite{Fu2021CVPR}, and SSHR~\cite{fu2023towards}. 
\textit{Non-referenced datasets} contain only raw or polarized RGB images without diffuse supervision; this category includes StereoMIS-Tracking~\cite{hayoz2023stereomis}, SCRREAM, HouseCat6D, CroMo, SCARED, and Cholec80. 
All endoscopic datasets belong to this group. To the best of our knowledge, no publicly available surgical dataset provides paired reflection-free ground truth.
\begin{table}
\centering
\scriptsize
\caption{Comparison of each method’s impact on pixel-matching, evaluated using the epipolar error ($E_{ep}$, lower is better $[\downarrow]$), and inlier ratio ($I_{R}$, higher is better $[\uparrow]$). Best results are in \textbf{bold}.}
\label{tab:downstream}
\begin{tabularx}{\linewidth}{@{}l|Y@{\hspace{1.5mm}}Y|Y@{\hspace{1.5mm}}Y|Y@{\hspace{1.5mm}}Y}

\midrule
\multicolumn{7}{c}{%
    {\fontsize{6.5pt}{7.5pt}\selectfont\textsc{\textbf{Natural Domain}}}
}\\
\midrule
  & \multicolumn{2}{c|}{CroMo} & \multicolumn{2}{c|}{HouseCat6D} &\multicolumn{2}{c}{SCRREAM} \\
\cmidrule(r){2-3} \cmidrule(r){4-5} \cmidrule(r){6-7}
 \textit{Method}& $E_{ep}$ $\downarrow$ & $I_{R}$ $\uparrow$ & $E_{ep}$ $\downarrow$ & $I_{R}$ $\uparrow$ & $E_{ep}$ $\downarrow$ & $I_{R}$ $\uparrow$ \\
\midrule
DHAN-SHR~\cite{guo2024dual}                        & 0.481           & 0.989           & \textbf{0.592}   & 0.995           & 0.831           & 0.997           \\
PA+PF~\cite{Yao2025CVPR}~\cite{Zhang2025ICCV}  & 0.364           & 0.992           & 0.722            & 0.998           & \textbf{0.369}  & 0.996           \\
SpecularityNet~\cite{Wu2022TMM}                    & 0.370           & 0.993           & 0.710            & 0.987           & 0.620           & 0.994           \\
Fu \textit{et al.}~\cite{fu2023towards}            & 0.532           & \textbf{0.997}  & 0.976            & \textbf{0.998}  & 0.386           & 0.997           \\
Li \textit{et al.}~\cite{li2025two}                & 0.467           & 0.992           & 0.606            & 0.997           & 0.578           & \textbf{0.999}  \\
EndoSTTN~\cite{daher2023temporal}                  & 0.326           & 0.996           & 0.952            & 0.996           & 0.867           & 0.999           \\
StableDelight~\cite{ye2024stablenormal}            & 0.370           & 0.993           & 1.042            & 0.994           & 0.599           & 0.991           \\
\midrule
OURS                                               & \textbf{0.301}  & \textbf{0.997}  & 0.759            & 0.991           & 0.423           & 0.998           \\
\midrule
\midrule
\multicolumn{7}{c}{%
    {\fontsize{6.5pt}{7.5pt}\selectfont\textsc{\textbf{Surgical Domain}}}
}\\
\midrule
\textit{}                     & \multicolumn{2}{c|}{Cholec80}      & \multicolumn{2}{c|}{SCARED}        & \multicolumn{2}{c}{StereoMIS-Tr.}   \\
\cmidrule(r){2-3} \cmidrule(r){4-5} \cmidrule(r){6-7}
\textit{Method}                                   & $E_{ep}$        & $I_{R}$         & $E_{ep}$         & $I_{R}$         & $E_{ep}$        & $I_{R}$         \\
\midrule
DHAN-SHR~\cite{guo2024dual}                       & 0.530           & \textbf{0.998}  & 1.013            & 0.996           & 0.729           & \textbf{1.000}           \\
PA+PF~\cite{Yao2025CVPR}~\cite{Zhang2025ICCV} & 0.567           & \textbf{0.998}  & 1.005            & 0.996           & 0.665           & \textbf{1.000}  \\
SpecularityNet~\cite{Wu2022TMM}                   & 0.587           & 0.995           & 1.155            & 0.996           & 0.213           & \textbf{1.000}           \\
Fu \textit{et al.}~\cite{fu2023towards}           & 0.437           & 0.997           & 1.155            & 0.997           & 0.958           & \textbf{1.000}           \\
Li \textit{et al.}~\cite{li2025two}               & 0.612           & 0.997           & 1.014            & 0.993           & 0.958           & \textbf{1.000}           \\
EndoSTTN~\cite{daher2023temporal}                 & 0.530           & 0.997           & \textbf{0.804}   & 0.988           & 1.922           & 0.974           \\
StableDelight~\cite{ye2024stablenormal}           & 0.409           & 0.996           & 0.898            & \textbf{0.998}  & 0.232           & 0.991           \\
\midrule
OURS                                              & \textbf{0.355}  & \textbf{0.998}  & 0.877            & \textbf{0.998}  & \textbf{0.167}  & \textbf{1.000}           \\
\bottomrule
\end{tabularx}
\end{table}

\subsection{Evaluation on Diffuse-Referenced Datasets}
\label{sec:eval_diffuse}
For datasets that provide paired diffuse ground truth, we directly evaluate the reconstructed diffuse image $\Idiff$ against its reference $\Idiff^\star$ using standard full-reference metrics (cf. Table~\ref{tab:withgt}), including mean squared error within the highlight masks ($\mathrm{MSE}_{m}$), Peak Signal-to-Noise Ratio (PSNR), and Structural Similarity Index Measure (SSIM). 
Across the PSD, SHIQ, and SSHR benchmarks, our method remains consistently competitive and provides clear advantages in terms of structural fidelity. 
On PSD, \textit{UnReflectAnything} delivers the strongest structural consistency, indicating that the model is particularly effective at restoring fine textures and attenuating residual highlight artifacts. 
On SHIQ and SSHR, while classical baselines perform with lower $\mathrm{MSE}_{m}$ and dominate in attenuating highlights, our approach yields the most stable structural outcomes (SSIM), which is notable given the severe saturation patterns and limited texture cues characteristic of this dataset. Our model attains lower $\mathrm{MSE}_{m}$ but also lower PSNR, indicating it may be less effective at preserving fine details in the non-highlighted regions despite achieving lower global error.

\begin{figure}
    \centering
    \includegraphics[width=\linewidth]{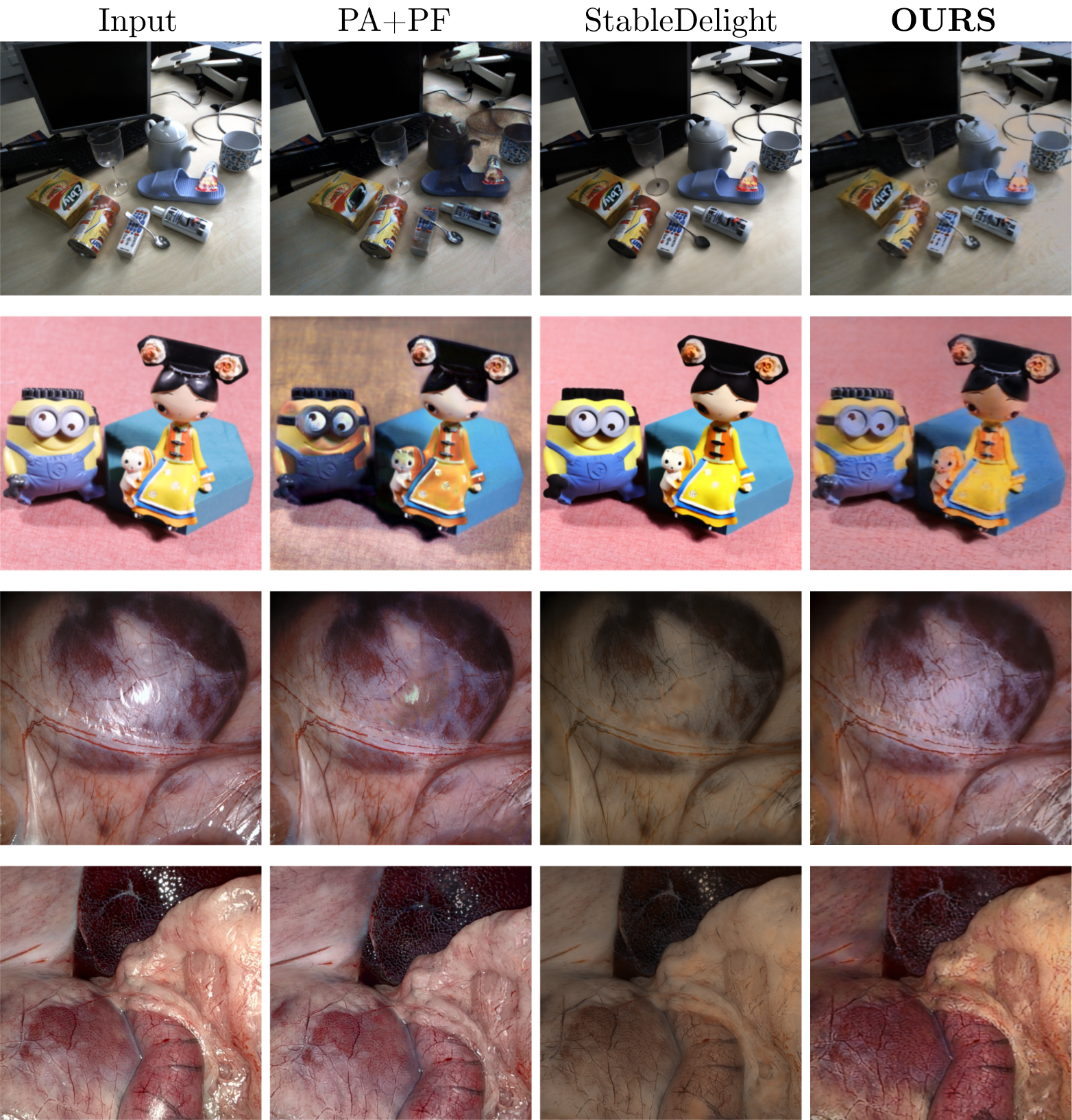}
\caption{\textbf{Qualitative comparison} between \textit{UnReflectAnything}, PolarAnything~\&~PolarFree (PA+PF), and StableDelight. OURS provides more consistent and effective attenuation of specular highlights across domains, while avoiding noticeable artefacts.}
    \label{fig:qualitativecomparison}
\end{figure}

\subsection{Evaluation on Non-Referenced Datasets}
\label{sec:eval_non-ref}
For datasets that lack diffuse ground truth, we evaluate (cf. Table~\ref{tab:nogt}) perceptual fidelity using 
NIQE~\cite{mittal2012no}, and a Luminance Suppression Ratio (LSR). 
NIQE assesses adherence to natural scene statistics, while we define LSR as a measure of how effectively highlight intensity is reduced without global dimming (for definition cf. Supp. Mat.).
Across the natural-domain datasets (CroMo, HouseCat6D, and SCRREAM), our method consistently achieves the strongest highlight suppression according to LSR, while maintaining NIQE scores that are competitive with or superior to those of existing baselines.
On surgical datasets, where specularities are more frequent and structured, 
our approach systematically attains the lowest LSR across Cholec80, SCARED, and StereoMIS-Tracking, indicating effective suppression of residual luminance in highlight regions, while remaining comparable to competing methods in terms of NIQE. 
These results suggests that the model can reduce specular artifacts without substantially degrading global perceptual quality.

\subsection{Downstream Performance}
We evaluate the influence of specular highlight suppression in a relative pose estimation task.
For each frame pair, we detect DISK~\cite{tyszkiewicz2020disk} keypoints, match them with LightGlue~\cite{lindenberger2023lightglue}, and estimate the essential matrix $\widehat{E}$ with MAGSAC++~\cite{barath2020magsac}.
We use symmetric epipolar error ($E_{ep}$) and inlier ratio ($I_R$) as comparison metrics (cf. Table~\ref{tab:downstream}).

Across the natural-domain datasets, our method yields competitive results while achieving state-of-the-art results on CroMo.
These findings indicate that the effect of highlight suppression varies across scenes, enhancing geometric alignment or correspondence stability depending on scene characteristics.
In surgical datasets, where highlights are more intense and widespread, our model shows a more uniform advantage.
It attains leading or joint-leading inlier ratios across Cholec80, SCARED, and StereoMIS-Tracking, and delivers competitive epipolar errors, including the lowest residuals on StereoMIS-Tracking.
These trends indicate reducing specular contamination helps preserve keypoint localization under challenging clinical illumination, where reflections often interfere directly with geometric cues.

The downstream results demonstrate suppressing specular peaks improves both the robustness and consistency of geometric estimation, supporting more reliable correspondence matching in both natural and surgical environments.

\subsection{Discussion}
\label{sec:discussion}
We present a diverse set of example predictions of \textit{UnReflectAnything} in Fig.~\ref{fig:qualitativeresults}, and a qualitative comparison with the most competitive SOTA methods in Fig.~\ref{fig:qualitativecomparison}.
Our framework consistently suppresses, inpaints, and attenuates specular highlights across diverse domains including challenging endoscopic imagery, where highlights are frequent and spatially complex. In these settings, diffusion-based baselines often exhibit incomplete inpainting, insufficient attenuation, or noticeable inconsistencies in local structure, whereas our method maintains coherent textures and geometry in the reconstructed regions.
While \textit{UnReflectAnything} demonstrates strong generalization across both natural and surgical domains, few limitations remain. The model tends to struggle with transparent or refractive objects, where the distinction between diffuse and specular components is inherently ambiguous. Our evaluations show sub-optimal performance in retaining structure and resolution in the non-highlight regions, which could be mitigated with integration of structure-aware priors, such as explicit edge constraints or high-frequency preservation modules. Incorporating stronger semantic reasoning could mitigate this limitation, as the current predictions rely predominantly on luminance cues. Additionally, we observe degraded reconstruction quality in cases of gradual, low-gradient highlights with smooth falloffs, where the inpainted seams may appear less coherent.

\section{Conclusion}
We present \textit{UnReflectAnything}, an RGB-only method for single-image highlight removal that learns from synthetically rendered supervision grounded in monocular geometry. By coupling a physically motivated highlight synthesis pipeline with token-space inpainting, our method achieves faithful diffuse reconstruction and generalizes across diverse real-world domains including challenging endoscopic imagery, without requiring paired data. \textit{UnReflectAnything} generalizes across diverse domains and achieves SOTA results on several benchmarks while maintaining strong overall performance, offering a solid foundation for RGB-only highlight removal with potential for clinical applications. Beyond visual improvement, \textit{UnReflectAnything} enhances geometric consistency in downstream correspondence and camera-pose estimation tasks, setting a new direction for physically guided learning in reflection suppression.



    

{
    \small
    \bibliographystyle{ieeetr}
    \bibliography{main}
}

\setcounter{section}{0}

\renewcommand*{\thesection}{\Alph{section}}
\renewcommand*{\thesubsection}{\Alph{subsection}}
\renewcommand*{\thesubsubsection}{\Alph{subsubsection}}

\clearpage
\setcounter{page}{1}
\maketitlesupplementary
\section{Extended Qualitative Inspection}
\label{supp:extendedqualitative}
We provide additional qualitative results of \textit{UnReflectAnything} beyond those shown in Fig.~\ref{fig:qualitativeresults}. These examples further illustrate the model's behaviour across diverse scenes and highlight its ability to suppress and inpaint specularities under varying appearance conditions.
\begin{figure}[b]
    \centering
    \includegraphics[width=\linewidth]{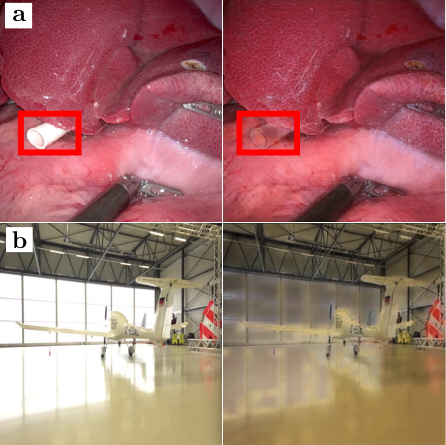}
    \caption{Representative failure modes of \textit{UnReflectAnything}.}
    \label{fig:failuremodes}
\end{figure}
\begin{figure*}
    \centering
    \includegraphics[width=0.92\linewidth]{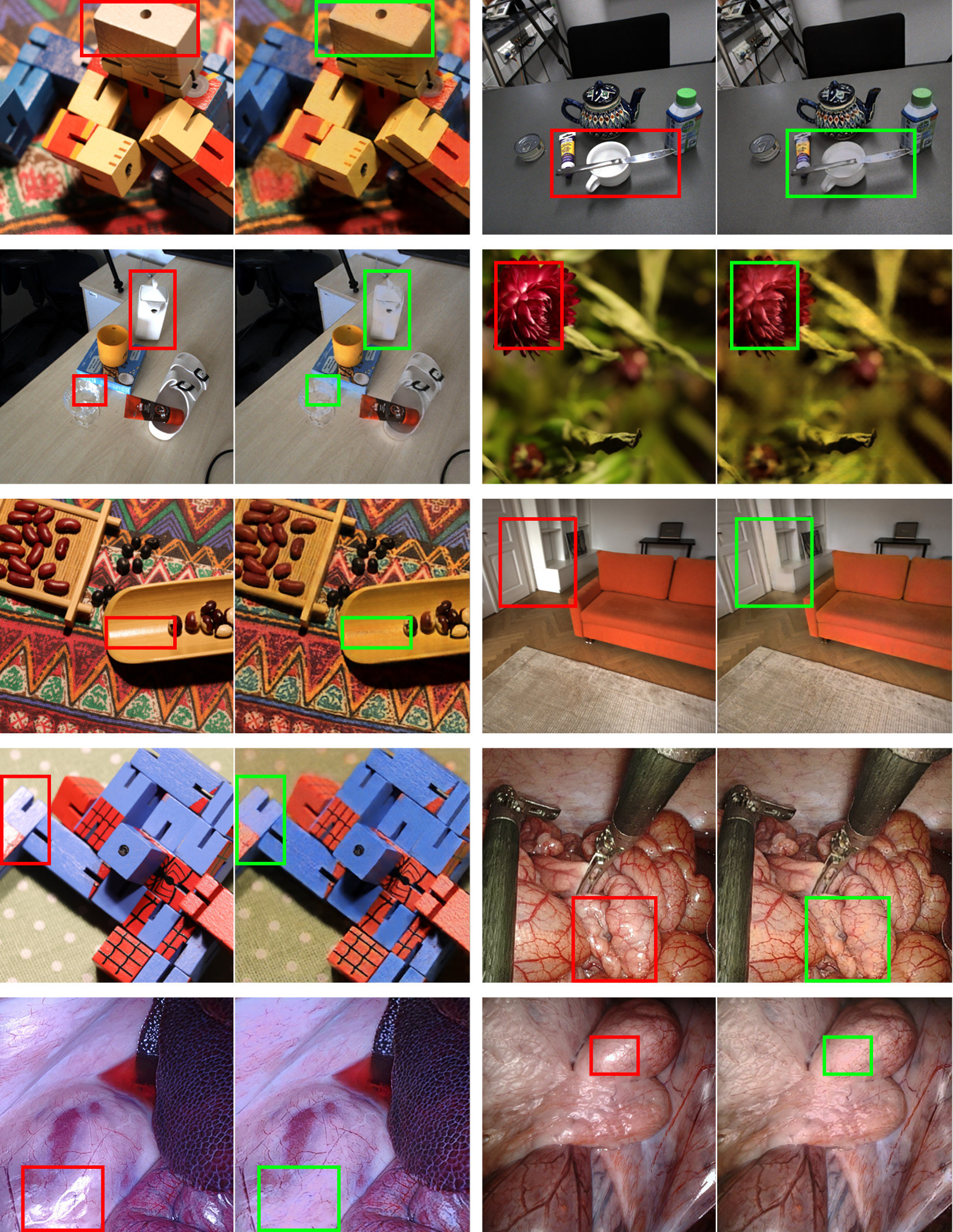}
    \caption{Additional input–output examples for \textit{UnReflectAnything} across multiple datasets. We indicate the presence of highlights in the input images (left of each pair) with a red rectangle and the highlight-free reconstruction in the output image (right of each pair) with a green rectangle.}
    \label{fig:extendedqualitative}
\end{figure*}

\section{Challenging Scenarios}

\label{sec:rationale}
As discussed in Sec.~\ref{sec:discussion}, although \textit{UnReflectAnything} generalizes well across multiple domains, some failure cases persist. We present illustrative examples in Fig.~\ref{fig:failuremodes}.  
A typical issue arises from the luminance-based detection of dataset highlights: highly reflective or bright surfaces, such as white anatomical structures in endoscopy, may be misinterpreted as specularities and subsequently inpainted (Fig.~\ref{fig:failuremodes}a).  
A similar phenomenon occurs in outdoor scenes (Fig.~\ref{fig:failuremodes}b), where very bright sky regions are incorrectly classified as highlights.

\section{Architecture Design Choices}
\label{supp:ablations}
We conduct additional ablations to further justify the architectural and supervisory choices in \textit{UnReflectAnything} (Table~\ref{tab:ablation}).
Replacing MoGe-2 normals with naïve depth-gradient normals markedly increases $MSE_m$, confirming that inaccurate normals produce unrealistic synthetic highlights and weaken supervision.
Disabling token-space inpainting and relying solely on RGB-space reconstruction (Eq.~\ref{eq:rgbloss}) produces blurrier outputs and reduces \textit{SSIM}, showing that restoring corrupted features before decoding is essential for preserving structure. Removing the local-mean prior further destabilizes token completion, particularly for large highlight regions.
Jointly training the decoder from scratch results in inferior performance compared to our two-stage curriculum, indicating that decoder pre-training offers a more stable feature-to-RGB initialization. Excluding dataset highlights from supervision is likewise crucial: supervising clipped pixels biases the model toward interpreting saturated reflections as diffuse regions, whereas masking them preserves a consistent and physically grounded learning signal.
Table~\ref{tab:lossweights} reports the numeric values of all loss weights used during training, which remain fixed throughout optimization.
\begin{table}[h!]
\centering
\footnotesize
\caption{\textbf{Ablation study} on supervision, model architecture, and training strategy. The reported $MSE_m$ and \textit{SSIM} metrics are averaged over the PSD, SSHR, SHIQ datasets.}
\label{tab:ablation}
\begin{tabularx}{\linewidth}{@{}l|Y|Y}
\toprule
\textit{Configuration}                             & \textit{$MSE_m$ $\downarrow$} & \textit{SSIM $\uparrow$} \\
\midrule
\textit{Supervision}                               &                  &               \\
\quad w/o MoGe-2 (depth-gradient normals)          &  0.012           &  0.909             \\
\midrule
\textit{Model Architecture}                        &                  &               \\
\quad w/o token inpainting (RGB inpainting)        &  0.007           &  0.816        \\
\quad w/o local mean prior                         &  0.004           &  0.911        \\
\midrule
\textit{Training Curriculum}                       &                  &               \\
\quad w/o decoder pre-training                     &  0.006           &  0.873        \\
\quad w/o dataset-highlight exclusion              &  0.022           &  0.933        \\
\midrule
\textbf{Full Model (OURS)}                         & \textbf{0.003}   & \textbf{0.957} \\
\bottomrule
\end{tabularx}
\end{table}

\label{supp:lossweights}
\begin{table}[h]
\centering
\footnotesize
\caption{Loss function weights used at training time.}
\label{tab:lossweights}
\begin{tabularx}{\linewidth}{@{}l|Y|Y@{}}
\toprule
\textbf{Loss Term} & \textbf{Symbol} & \textbf{Value} \\
\midrule
Highlight Dice loss            & $w_{\mathrm{dice}}$ & $0.2$ \\
Highlight L1 loss              & $w_{\mathrm{L1}}$   & $0.7$ \\
Highlight TV regularizer       & $w_{\mathrm{TV}}$   & $0.1$ \\
Seam loss                      & $w_{\mathrm{seam}}$ & $0.25$ \\
Specularity penalty            & $w_{\mathrm{spec}}$ & $0.25$ \\
Diffuse RGB reconstruction     & $w_{\mathrm{RGB}}$  & $0.5$ \\
\bottomrule
\end{tabularx}
\end{table}


\end{document}